\documentclass[10pt,twocolumn,letterpaper]{article}

\usepackage{iccv}
\usepackage{times}
\usepackage{epsfig}
\usepackage{graphicx}
\usepackage{amsmath}
\usepackage{amssymb}

\usepackage{bm}
\usepackage{amsmath, amsfonts, amsthm}
\usepackage{float}
\usepackage[ruled,vlined,linesnumbered]{algorithm2e}
\usepackage{algpseudocode}
\usepackage{leftidx}
\usepackage{url}
\usepackage{graphicx}
\usepackage{tabularx, multirow, multicol, booktabs}
\usepackage{color}
\usepackage{soul}
\usepackage{subfigure}
\usepackage{caption}
\usepackage{float}
\usepackage{balance}

\usepackage{pifont}
\newcommand{\cmark}{\ding{51}}%
\newcommand{\xmark}{\ding{55}}%

\newcommand{\vae}{VAE\xspace}

\newcommand{\gru}{GRU\xspace}
\newcommand{\lstm}{LSTM\xspace}

\newcommand{\dtw}{DTW\xspace}
\newcommand{\dicnn}{DiCNN\xspace}

\newcommand{\proposed}{WETAS\xspace}
\newcommand{\donut}{Donut\xspace}
\newcommand{\lstmvae}{LSTM-VAE\xspace}
\newcommand{\lstmndt}{LSTM-NDT\xspace}
\newcommand{\omnianom}{OmniAnomaly\xspace}

\newcommand{\deepmil}{DeepMIL\xspace}

\newcommand{\donutpp}{Donut\textsuperscript{++}\xspace}
\newcommand{\lstmvaepp}{LSTM-VAE\textsuperscript{++}\xspace}
\newcommand{\lstmndtpp}{LSTM-NDT\textsuperscript{++}\xspace}
\newcommand{\omnianompp}{OmniAnomaly\textsuperscript{++}\xspace}

\newcommand{\smd}{SMD\xspace}

\newcommand{\ghl}{GHL\xspace}

\newcommand{\emg}{EMG\xspace}
\newcommand{\subway}{Subway\xspace}

\newcommand{\subwayexit}{Subway\xspace}

\newcommand{\vect}{\mathbf{X}}
\newcommand{\vech}[1]{\mathbf{h}_{#1}}
\newcommand{\vecx}[1]{\mathbf{x}_{#1}}
\newcommand{\weightg}{\mathbf{w}}
\newcommand{\weightl}{\mathbf{w}}
\newcommand{\actmap}[1]{m_{#1}}
\newcommand{\actmaps}{\mathbf{m}}
\newcommand{\score}[1]{s_{#1}}
\newcommand{\scores}{\mathbf{s}}

\newcommand{\wlab}{y}
\newcommand{\slab}[1]{\widetilde{z}_{#1}}
\newcommand{\slabs}{\widetilde{\mathbf{z}}}
\newcommand{\aslab}[1]{z_{#1}}
\newcommand{\aslabs}{\mathbf{z}}
\newcommand{\sdtw}[1]{\text{DTW}_{#1}}

\newcommand{\sumfuc}{\bm{\phi}_L}

\newcommand{\closs}{\mathcal{L}_{c}}
\newcommand{\aloss}{\mathcal{L}_{a}}

\newcommand{\smallsection}[1]{{\vspace{0.05in} \noindent \bf {#1.\hspace{5pt}}}}
\DeclareMathOperator*{\argmin}{argmin}

\SetCommentSty{mycommfont}

\let\oldnl\nl
\newcommand{\nonl}{\renewcommand{\nl}{\let\nl\oldnl}}

\newcolumntype{P}{>{\centering\arraybackslash}m{0.125\linewidth}}
\newcolumntype{Q}{>{\centering\arraybackslash}m{0.09\linewidth}}
\newcolumntype{A}{>{\raggedleft\arraybackslash}m{0.31\linewidth}}
\newcolumntype{B}{>{\centering\arraybackslash}m{0.14\linewidth}}
\newcolumntype{H}{>{\centering\arraybackslash}m{0.11\linewidth}}

\usepackage[pagebackref=true,breaklinks=true,letterpaper=true,colorlinks,bookmarks=false]{hyperref}

\iccvfinalcopy 

\def\iccvPaperID{5792} 

\ificcvfinal\fi

\begin{document}

\title{Weakly Supervised Temporal Anomaly Segmentation with~Dynamic~Time~Warping}

\author{Dongha Lee\textsuperscript{1}\thanks{This work was done when the author was at POSTECH.} , Sehun Yu\textsuperscript{2}, Hyunjun Ju\textsuperscript{2}, Hwanjo Yu\textsuperscript{2}\thanks{Corresponding author.}\\
\textsuperscript{1}University of Illinois at Urbana-Champaign (UIUC), Urbana, IL, United States\\
\textsuperscript{2}Pohang University of Science and Technology (POSTECH), Pohang, South Korea\\
{\tt\small donghal@illinois.edu, \{hunu12,hyunjunju,hwanjoyu\}@postech.ac.kr}
}

\maketitle
\ificcvfinal\thispagestyle{empty}\fi

\begin{abstract}
Most recent studies on detecting and localizing temporal anomalies have mainly employed deep neural networks to learn the normal patterns of temporal data in an unsupervised manner.
Unlike them, the goal of our work is to fully utilize instance-level (or weak) anomaly labels, which only indicate whether any anomalous events occurred or not in each instance of temporal data.
In this paper, we present \proposed, a novel framework that effectively identifies anomalous temporal segments (i.e., consecutive time points) in an input instance.
\proposed learns discriminative features from the instance-level labels so that it infers the sequential order of normal and anomalous segments within each instance, which can be used as a rough segmentation mask.
Based on the dynamic time warping (DTW) alignment between the input instance and its segmentation mask, \proposed obtains the result of temporal segmentation, and simultaneously, it further enhances itself by using the mask as additional supervision.
Our experiments show that \proposed considerably outperforms other baselines in terms of the localization of temporal anomalies, and also it provides more informative results than point-level detection methods.
\end{abstract}

\section{Introduction}
\label{sec:intro}
Anomaly detection, which refers to the task of identifying anomalous (or unusual) patterns in data, has been extensively researched in a wide range of domains, such as fraud detection~\cite{abdallah2016fraud}, network intrusion detection~\cite{ju2020pumad}, and medical diagnosis~\cite{wei2018anomaly}.
In particular, detecting the anomaly from temporal data (e.g., multivariate time-series and videos) has gained much attention in many real-world applications, for finding out anomalous events that resulted in unexpected changes of a temporal pattern or context.

Recently, several studies on anomaly detection started to localize and segment the anomalies within an input instance based on deep neural networks~\cite{bergmann2019mvtec, bergmann2020uninformed, hendrycks2019benchmark}, unlike conventional methods which simply classify each input instance as positive (i.e., anomalous) or negative (i.e., normal). 
In this work, we aim to precisely localize the anomalies in temporal data by detecting anomalous temporal segments, defined as the group of consecutive time points relevant to anomalous events.
Note that labeling every anomalous time point is neither practical nor precise, similarly to other segmentation problems~\cite{chang2019d3tw, lee2019ficklenet, richard2018neuralnetwork}.
In this sense, the main challenge of temporal anomaly segmentation is to distinguish anomalous time points from normal time points without using point-level anomaly labels for model training.

\begin{figure}[t]
	\centering
	\includegraphics[width=\linewidth]{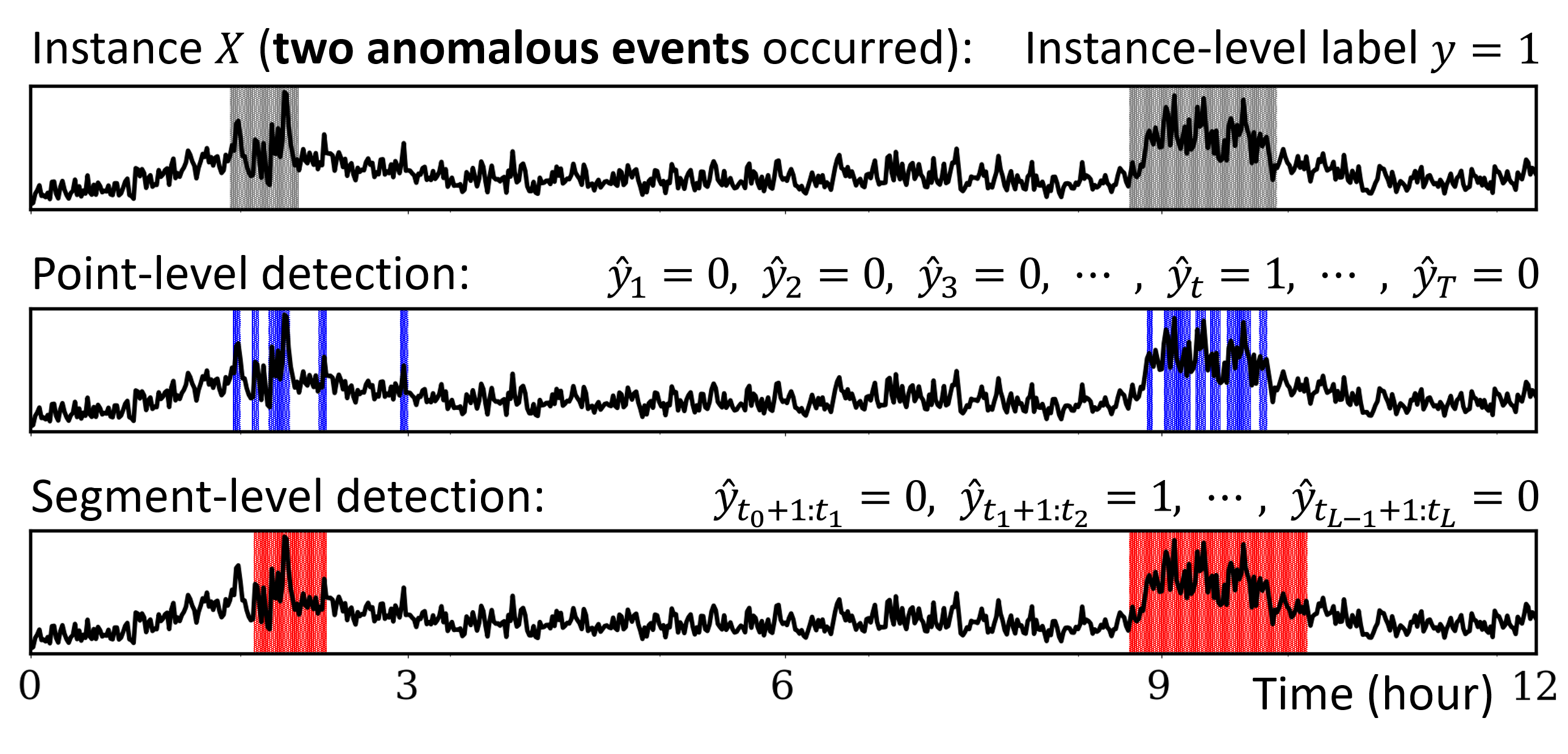}
	\vspace{-15pt}
	\caption{Two different strategies for localizing temporal anomalies. Given an input instance of length $T$, the point-level and segment-level anomaly detection produce the series of binary labels of length $T$ and length $L$, respectively.}
	\label{fig:detection}
\end{figure}

The most dominant approach to the localization of the temporal anomalies is point-level anomaly detection based on the reconstruction of input instances~\cite{ahmad2017unsupervised,hasan2016learning,hundman2018detecting,luo2017remembering,munir2018deepant,park2018multimodal,su2019robust,xu2018unsupervised}.
To learn the point-level (or local) anomaly score in an unsupervised manner, they focus on modeling the \textit{normal} patterns (i.e., temporal context of each time point, usually given in a form of its past inputs~\cite{chandola2009anomaly}) from \textit{normal} data, while considering all unlabeled data as normal.
Specifically, they mainly employ variational autoencoder (\vae) to learn normal latent vectors of temporal inputs, or train sequence models to predict the next input by using RNNs or CNNs.
Then, they compute the anomaly score for each time point based on the magnitude of error between an actual input and the reconstructed (or predicted) one.

However, such unsupervised learning methods show limited performance due to the absence of information about the target anomalies. %
They are not good at discovering specific patterns caused by an anomalous event, especially in the case that such patterns reside in training data.
As a solution, we argue that instance-level labels are often easy to acquire by simply indicating the occurrence of any anomalous events, whereas acquiring point-level labels is hard in practice.
For example, given a time-series instance collected for a fixed period of time, a human annotator can easily figure out whether any anomalous events occurred or not during the period, then generate a binary label for the instance.

In this paper, we propose a novel deep learning framework, named as \proposed, which leverages \underline{WE}ak supervision for \underline{T}emporal \underline{A}nomaly \underline{S}egmentation.
\proposed learns from labeled ``instances'' of temporal data in order to detect anomalous ``segments'' within the instance.
The segment-level anomaly detection is more realistic than the point-level detection (Figure~\ref{fig:detection}), because anomalous events usually result in variable-length anomalous segments in temporal data.
The point-level detection cannot tell whether two nearby points detected as the anomaly come from a single anomalous event or not, and thus the continuity of detected points should be investigated for further interpretation of the results.
Our problem setting is similar to the multiple-instance learning (MIL)~\cite{sultani2018real} which learns from bag-level labels,\footnote{The term ``multiple-instance'' used in MIL refers to same-length temporal segments that compose a single bag. Thus, ``instance'' and ``bag'' in MIL respectively correspond to ``segment'' and ``instance'' in our approach.} but differs in that MIL makes predictions for \textit{static} segments of the same length.
Table~\ref{tbl:comparison_tad} summarizes recent anomaly detection methods for temporal data.

\renewcommand{\arraystretch}{0.9}
\begin{table}[t]
    \centering
    \small
    \resizebox{0.99\linewidth}{!}{%
    \begin{tabular}{rrc}
        \toprule
        \multicolumn{1}{c}{Methods} & \multicolumn{1}{c}{Anomaly Prediction} & Instance Label \\\midrule
        Park et al.~\cite{park2018multimodal} & point-level & \xmark \\
        Su et al.~\cite{su2019robust} & point-level & \xmark \\ 
        Xu et al.~\cite{xu2018unsupervised} & point-level &  \xmark \\
        Sultani et al.~\cite{sultani2018real} & (\textbf{static}) segment-level & \cmark \\
        {\proposed (ours)} & (\textbf{dynamic}) segment-level & \cmark \\\bottomrule
    \end{tabular}
    }
    \caption{Comparison of anomaly detection methods that are able to localize the temporal anomalies in each instance.}
    \label{tbl:comparison_tad}
\end{table}

\proposed fully exploits the instance-level (or weak) labels for training its model and also for inferring the sequential anomaly labels (i.e., pseudo-labeling) that can be used as rough segmentation masks.
Based on dynamic time warping (\dtw) alignment~\cite{sakoe1978dynamic} between an input instance and its sequential pseudo-label, \proposed effectively finds variable-length anomalous segments.
To be specific, \proposed optimizes the model to accurately classify an input instance as its instance-level label, and simultaneously to best align the instance with its sequential pseudo-label based on the DTW.
As the training progresses, the model generates more accurate pseudo-labels, and this eventually improves the model itself by the guidance for better alignment between inputs and their pseudo-labels.

Our extensive experiments on real-world datasets, including multivariate time-series and surveillance videos, demonstrate that \proposed successfully learns the normal and anomalous patterns under the weak supervision.
Among various types of baselines, \proposed achieves the best performance in detecting anomalous points.
Furthermore, the qualitative comparison of the detection results shows that point-level detection methods identify non-consecutive anomalous points even for a single event, whereas our framework obtains the smoothed detection results which specify the starting and end points of each event.

\section{Related Work}
\label{sec:related}

\subsection{Anomaly Detection for Temporal Data}
\label{subsec:related_ad}
Most recent studies on anomaly detection for temporal data (e.g., multivariate time-series) utilize deep neural networks to learn the temporal normality (or regularity) from training data~\cite{ahmad2017unsupervised,hundman2018detecting,munir2018deepant}.
In particular, VAE-based models~\cite{hasan2016learning,luo2017remembering,park2018multimodal,su2019robust,xu2018unsupervised} have gained much attention because of their capability to capture the normal patterns in an unsupervised manner.
Their high-level idea is to compute the point-level anomaly scores by measuring how well a temporal context for each time point (i.e., a sliding window-based temporal input) can be reconstructed using the VAE.
They are able to localize the temporal anomalies within an input instance to some extent by the help of this point-level detection (or alert) approach.
However, they do not utilize information about the anomaly (e.g., anomaly labels) for model training, which makes it difficult to accurately identify the anomalies.
In general, as the anomalies are not limited to simple outliers or extreme values, modeling the \textit{abnormality} as well is helpful to learn useful features for discrimination between normal and anomalous time points~\cite{gornitz2013toward, pang2019deep}.


Several studies make use of anomaly labels by adopting U-Net architectures~\cite{wen2019time,zhang2018human} which are known to be effective for spatial segmentation~\cite{long2015fully,ronneberger2015u}.
However, their models need to be trained with full supervision, which means that the model training is guided by anomaly labels for every time point.
It makes the detector impractical because labeling every point in each instance or obtaining such point-level labels is infeasible or costs too much in practice.

\subsection{Weakly Supervised Temporal Segmentation}
\label{subsec:related_weakseg}
To address the limitation of fully supervised temporal segmentation, weakly supervised approaches have been actively researched for video action segmentation (or detection) tasks~\cite{chang2019d3tw, richard2018neuralnetwork, yu2019temporal, yuan2017temporal}.
They aim to learn weakly annotated data whose labels are not given for every time point (frame) in an instance (video).
To this end, dynamic program-based approaches are employed to search for the best results of segmentation over the time axis under the weak supervision, including the Viterbi algorithm~\cite{richard2018neuralnetwork} and dynamic time warping~\cite{chang2019d3tw, yu2019temporal, yuan2017temporal}. 
Nevertheless, they cannot be directly applied to anomaly segmentation, because they require instance-level \textit{sequential} labels being used as rough segmentation masks, or they impose strong constraints on the occurrence of target events (actions).\footnote{They assume that each action occurs following a specific transition diagram~\cite{yu2019temporal} or occurs at most once in a single instance~\cite{yuan2017temporal}.}

Without the help of sequential labels and constraints, the multiple-instance learning (MIL) approach~\cite{sultani2018real} showed promising results in detecting anomalous events by leveraging weakly labeled videos.
It considers each video as a bag after dividing the video into a fixed number of same-length segments, then trains an anomaly detection model by using both positive (i.e., anomalous) and negative (i.e., normal) bags.
The trained model is able to predict the label of each video segment as well as a bag, thus the results can be utilized for anomaly segmentation.
However, due to its statically-segmented inputs, there exists a trade-off between accurate anomaly detection and precise segmentation with respect to the number of segments in a single bag.
The more (and shorter) segments make it harder to capture the long-term contexts within the segments, whereas the fewer (and longer) segments generate coarse-grained segmentation results which could much differ from the actual observation.

\section{Temporal Anomaly Segmentation}
\label{sec:proposed}

\subsection{Problem Formulation}
\label{subsubsec:problem}
The goal of temporal anomaly segmentation is to specify variable-length anomalous segments (i.e., the starting and end points) within an input instance of temporal data.
Formally, given a $D$-dimensional input instance $\vect = [\vecx{1}, \vecx{2}, \ldots, \vecx{T}] \in \mathbb{R}^{D\times T}$ of temporal length $T$, we aim to 
produce segment-level anomaly predictions $\hat{\mathbf{y}}=[\hat{y}_{t_0+1:t_1}, \hat{y}_{t_1+1:t_2}, \ldots, \hat{y}_{t_{L-1}+1:t_L}] \in \{0,1\}^{L}$ where $t_l$ denotes the end point of the $l$-th segment (i.e., $t_0=0$ and $t_L=T$).
This task can be treated similarly to the point-level anomaly detection from the perspective of its final output $\hat{\mathbf{y}}$, but differs in that it guarantees a one-to-one mapping between identified segments and anomalous events.

In this work, we focus on \textit{weakly supervised learning}, in order to bypass the challenge of obtaining densely labeled temporal data for model training.
Specifically, a training set consists of $N$ instances\footnote{Each instance is collected for the fixed period of time $T$, or obtained by splitting data streams into fixed-length temporal data.} with their instance-level binary labels, $\{(\vect^{(1)}, y^{(1)}), (\vect^{(2)}, y^{(2)}), \ldots, (\vect^{(N)}, y^{(N)})\}$.
In this context, the weak supervision means that each label simply indicates whether any anomalous events are observed or not in the instance.
Only with the weakly annotated dataset, we train the model for temporal anomaly segmentation, where temporal data are given without any labels at test time.

\subsection{Dynamic Time Warping (DTW) Alignment for Temporal Anomaly Segmentation}
\label{subsec:dtwseg}
We first present how to obtain the segmentation result of an input instance by using a rough segmentation mask.
To this end, we define an additional type of anomaly labels which can serve as the segmentation mask, referred to as sequential anomaly label.\footnote{The details about synthesizing this label are discussed in Section~\ref{subsec:weaklearn}.}
This label indicates the existence (and order) of normal and anomalous events for each input instance; $\aslabs=[\aslab{1},\aslab{2},\ldots,\aslab{L}]\in\{0, 1\}^L$ where $L$ is the length of the sequential label.
For example, $\aslabs = [0, 1, 0, 1, 0]$ in case of $L=5$ means that two anomalous events are observed within the instance.
Note that the sequential label $\aslabs$ itself cannot be the final output of our framework, because it does not contain the information about the starting and end points of each segment.

For temporal anomaly segmentation, we utilize dynamic time warping (DTW)~\cite{sakoe1978dynamic} which outputs the optimal alignment between a target instance and the sequential anomaly label.
The goal of DTW is to find the optimal alignment (i.e., point-to-point matching) between $\aslabs$ and $\vect$, which minimizes their total alignment cost with the time consistency among the aligned point pairs.
To be precise, the total alignment cost is defined by the inner product of the cost matrix $\Delta(\aslabs, \vect) \in \mathbb{R}^{L \times T}$ and the binary alignment matrix $A\in\{0,1\}^{L \times T}$.
Each entry of the cost matrix, $[\Delta(\aslabs, \vect)]_{lt} := \delta(\aslab{l}, \vecx{t})$, encodes the cost for aligning $\vecx{t}$ with $\aslab{l}$ (i.e., the penalty for labeling $\vecx{t}$ as $\aslab{l}$), and similarly,
the entry of $A$ indicates the alignment between $\aslabs$ and $\vect$;
that is, $A_{lt}=1$ if $\vecx{t}$ is aligned with $\aslab{l}$ and $A_{lt}=0$ otherwise.
The optimal alignment matrix $A^*$ is obtained by
\begin{equation}
\small
    A^* = \argmin_{A\in\mathcal{A}}\ \langle A, \Delta(\aslabs, \vect)\rangle,
\end{equation}
where $\mathcal{A}$ is the set of possible binary alignment matrices.

\begin{figure*}[t]
	\centering
	\includegraphics[width=\linewidth]{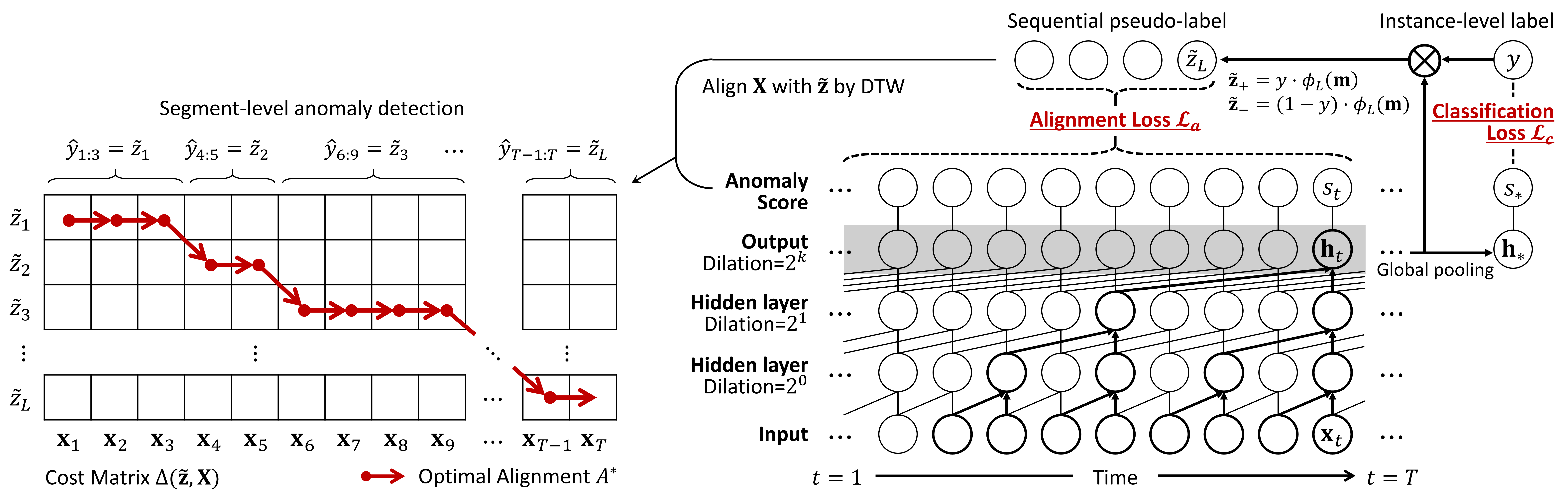}
	\caption{The overall framework of \proposed, optimized by both the classification loss $\closs$ and alignment loss $\aloss$. The red arrows in the cost matrix $\Delta(\slabs, \vect)$ are the desired alignment between a temporal input (column) and its sequential label (row) with time consistencies. Based on the optimal alignment, \proposed can produce the segment-level anomaly predictions.}
	\label{fig:framework}
\end{figure*}

In order to align the two series ($\aslabs$ and $\vect$) in a temporal order, each alignment matrix is enforced to represent a single path on a $L \times T$ matrix that connects the upper-left $(1, 1)$-th entry to the lower-right $(L, T)$-th entry using $\rightarrow, \searrow$ moves.
Based on the obtained alignment $A^*$, the starting and end points of the $l$-th temporal segment and its anomaly label are determined by $\hat{y}_{t_{l-1}+1:t_l}=\aslab{l}$ such that $A^*_{lt}=1$ for all $t_{l-1}< t\leq t_l$.
For example, in Figure~\ref{fig:framework} (Left), the red arrows represent non-zero entries in the optimal alignment matrix, and we can easily identify the segments whose segment-label is 1 (i.e., $z_l=1$) as the anomaly. 
Unlike the conventional \dtw, $\downarrow$ moves are not available in our warping paths because each time point should be aligned with only a single label.


The two key challenges are raised here as follows.
1) We need to carefully model the cost matrix based on the anomaly score so that it can effectively capture the \textit{normality} and \textit{abnormality} at each point, while considering the long-term dependency. 
In addition, 2) we need to obtain sequential labels used for the DTW alignment by distilling the information from given instance-level labels.

\subsection{Anomaly Score Modeling for Cost Matrix}
\label{subsec:model}
To model the cost function $\delta(\aslab{l}, \vecx{t})$ that is required to define the cost matrix $\Delta(\aslabs, \vect)$, we parameterize the anomaly score at time $t$ (i.e., the probability that $\vecx{t}$ comes from an anomalous event) by using deep neural networks.
Among various types of autoregressive networks designed for temporal data, we employ the basic architecture of dilated CNN (\dicnn) introduced by WaveNet~\cite{oord2016wavenet}.
As illustrated in Figure~\ref{fig:framework} (Right), \dicnn is basically stacks of dilated \textit{causal} convolutions, which is the convolution operation not considering future inputs when computing the output at each point.
Specifically, the $n$-th layer has a convolution with filter size $k$ and dilation rate $k^{n-1}$, thus 
the output vector from layer $n$ at time $t$ (denoted by $\vech{t}$) depends on its previous $k^n$ points in the end;
i.e., $[\vecx{t-k^n+1}, \ldots, \vecx{t}]$ becomes its receptive field.


Using the inner product of the output vector at each point $\vech{t}\in\mathbb{R}^{d}$ and the anomaly weight vector $\weightl\in\mathbb{R}^{d}$, whose parameters are trainable, we compute the series of local anomaly scores $\scores \in \mathbb{R}^T$ by
\begin{equation}
\small
    \scores = [\score{1}, \score{2}, \ldots, \score{T}], \text{\ where\ \  } \score{t} = \sigma(\weightl{}^\top\vech{t}).
\label{eq:lscore}
\end{equation}
We apply the sigmoid function $\sigma$ to make the score range in $[0, 1]$.
Finally, we define the cost $\delta(\aslab{l}, \vecx{t})$ by the negative log-posterior probability using the anomaly score $\score{t}$:
\begin{equation}
\small
\label{eq:distfunc}
\begin{split}
\delta(\aslab{l}, \vecx{t})
&= -\log p \left(\aslab{l}|\vecx{t} \right) \\
&= -\left\{\aslab{l}\cdot\log\score{t} + (1-\aslab{l})\cdot\log(1-\score{t}) \right\}.
\end{split}
\end{equation}

\noindent
The cost becomes large when the computed score $\score{t}$ is far from a target label $\aslab{l}$.
In this sense, DTW can identify the anomalous segments by finding the optimal point-to-point matching between the series of anomaly scores and its sequential label, which minimizes the total alignment cost.

\subsection{Learning with Weak Supervision}
\label{subsec:weaklearn}
Our proposed framework, termed as \proposed, optimizes the \dicnn model by leveraging only weak supervision for the anomaly segmentation.
To fully utilize the given instance-level labels, two different types of losses are considered: 
1) the classification loss for correctly classifying an input instance as its instance-level anomaly label, and 2) the alignment loss for matching the input instance well with the sequential label, which is synthesized by the model by distilling the instance-level label.

\smallsection{Learning from Instance-level Label}
\label{subsubsec:learn_weaklabel}
Similar to the \textit{local} (or point-level) anomaly score in Equation~\eqref{eq:lscore}, we define the \textit{global} (or instance-level) anomaly score for its binary classification.
The global anomaly score $\score{*}\in\mathbb{R}$ is computed by using the anomaly weight $\weightg\in\mathbb{R}^d$ and the global output vector $\vech{*}\in\mathbb{R}^d$, obtained by global max (or average) pooling on all the output vectors along the time axis,
\begin{equation}
\small
\begin{split}
    \label{eq:gmp}
    \vech{*} &= \text{global-pooling}\ (\vech{1}, \vech{2}, \ldots, \vech{T}) \\
    \score{*} &= \sigma(\weightg{}^\top\vech{*}).
\end{split}
\end{equation}

To optimize our model based on the instance-level labels, we define the classification loss by the binary cross entropy between $\wlab$ and $\score{*}$.
Thereby, we can predict the anomaly label for an input instance based on its global anomaly score:
\begin{equation}
\small
\label{eq:closs}
        \closs 
        = -\sum_{i=1}^{N}\left\{\wlab^{(i)} \cdot \log \score{*}^{(i)} + (1-\wlab^{(i)})\cdot\log(1-\score{*}^{(i)})\right\}.
\end{equation}
The classification loss optimizes the model to differentiate between normal patterns and anomalous patterns that are commonly observed in weakly labeled temporal data.

\smallsection{Learning from Sequential Pseudo-Label}
\label{subsubsec:learn_seqlabel}
To begin with, we propose the technique to infer the sequential anomaly label (i.e., pseudo-label generation) of an input instance, then utilize the generated sequential pseudo-label to further improve our \dicnn model so that it can compute more accurate local anomaly scores.
The pseudo-labeling technique is motivated by temporal class activation map (CAM)~\cite{lee2021learnable, wang2017time} which helps to analyze temporal regions that most influence the instance-level classification.
We obtain the anomaly activation map $\actmaps=[\actmap{1}, \actmap{2}, \ldots, \actmap{T}] \in \mathbb{R}^{T}$ by multiplicating each output vector $\vech{t}$ with the weight vector $\weightg$ that is used for the instance-level anomaly classification.
That is, $\actmap{t}$ becomes proportional to how strongly the time point $t$ contributes to classifying the input instance as the anomaly.
For the condition $0\leq\actmap{t}\leq1$, their values are min-max normalized along the time axis within each instance,
\begin{equation}
\small
\label{eq:actmap}
    \actmap{t} = \frac{\weightg^\top\vech{t} - \min_{t'} \weightg^\top\vech{t'}}{\max_{t'} \weightg^\top\vech{t'} - \min_{t'} \weightg^\top\vech{t'}}.
\end{equation}

Then, we introduce a pseudo-labeling function $\sumfuc:\mathbb{R}^T \rightarrow \{0, 1\}^L$ which partitions a whole anomaly activation map into $L$ segments (or disjoint intervals) of the same length $\lceil T/L \rceil$ and generates a binary label for each segment, i.e., $\slabs =\sumfuc(\actmaps)\in\{0, 1\}^L$. 
The pseudo-label $\slab{l}$ is determined by whether the maximum activation value in the $l$-th segment is larger than the anomaly threshold $\tau$ or not:
\begin{equation}
\small
\label{eq:seqlabel}
    \slab{l} = \mathbb{I}\left[\left(\max_{(l-1) \cdot \lceil T/L \rceil \leq t \leq l \cdot \lceil T/L \rceil} \actmap{t}\right) \geq \tau \right].
\end{equation}
This sequential pseudo-label $\slabs$ offers information on how many anomalous segments do exist and also their relative locations.
Note that it is used as the rough segmentation mask for the DTW alignment, explained in Section~\ref{subsec:dtwseg}.


The alignment loss is designed to reduce the DTW discrepancy between an input $\vect$ and its sequential pseudo-label $\slabs$ for their better alignment.
However, the gradient of DTW is not very well defined (i.e., not differentiable) with respect to the cost matrix, due to its discontinuous hard-min operation taking only the minimum value.
Thus, we instead adopt the soft-DTW~\cite{cuturi2017soft} that provides the continuous relaxation of DTW by combining it with global alignment kernels~\cite{cuturi2007kernel}.
The soft-DTW utilizes the soft-min operation,
\begin{equation}
\small
\label{eq:softdtw}
    \sdtw{\gamma}(\slabs, \vect) =  \text{min}_{\gamma}\left\{ \langle A, \Delta(\slabs, \vect)\rangle, \forall A\in \mathcal{A} \right\}
\end{equation}
where $\text{min}_{\gamma}{}$ 
with a smoothing parameter $\gamma$ is defined by 
\begin{equation*}
\small
\label{eq:softmin}
    \text{min}_{\gamma}\{a_1, a_2, \ldots, a_n\} =
    \begin{cases}
    \; \min_{i\leq n} a_i, &\gamma = 0\\
    \; -\gamma \log \sum_{i=1}^{n}e^{-a_i/\gamma}, &\gamma > 0.
    \end{cases}
\end{equation*}
The soft-DTW distance in Equation~\eqref{eq:softdtw} can be obtained by solving a dynamic program based on Bellman's recursions. 
Please refer to~\cite{cuturi2017soft, lee2021learnable} for its forward and backward recursions to compute $\sdtw{\gamma}(\slabs, \vect)$ and $\nabla_{\Delta(\slabs, \vect)}\sdtw{\gamma}(\slabs, \vect)$.

We remark that only minimizing $\sdtw{\gamma}(\slabs, \vect)$ causes the problem of \textit{degenerated alignment}~\cite{chang2019d3tw} that assigns each label to a single time point.
To obtain more precise boundaries of segmentation results, we use the discriminative modeling under the supervision of the obtained pseudo-label.
In detail, we optimize the model so that the alignment with a positive pseudo-label $\slabs_+$ costs less than that with a negative pseudo-label $\slabs_-$ by a margin $\beta$ as follows.
\begin{equation}
\small
\label{eq:aloss}
    \aloss = 
    \sum_{i=1}^{N} \left[\frac{1}{T}\sdtw{\gamma}(\slabs_+^{(i)}, \vect^{(i)})- \frac{1}{T}\sdtw{\gamma}(\slabs_-^{(i)}, \vect^{(i)})  +\beta \right]_{+},
\end{equation}
where $[z]_{+} = \max(z, 0)$.
The positive and negative pseudo-labels are respectively obtained by $\slabs_+ = \wlab\cdot\sumfuc(\actmaps)$ and $\slabs_- =(1-\wlab)\cdot\sumfuc(\actmaps)$, thereby the instance-level label $\wlab$ plays the role of a binary mask.
In other words, we regard the sequential pseudo-label corrupted by the wrong instance-level label (i.e., $1-\wlab$) as negative.
This loss is helpful to produce more accurate local anomaly scores
by making them align better with its sequential pseudo-label.
Consequently, the model is capable of exploiting richer supervision than only using the instance-level labels.

\subsection{Optimization and Inference}
\label{subsec:optimization}
The final loss is described by the sum of the two losses, $\mathcal{L}=\closs+\aloss$, where we can control the importance of the alignment loss by adjusting the margin size $\beta$.
Note that the anomaly weight vector $\weightg$ and the network parameters in the \dicnn model are effectively optimized by minimizing the final loss, whereas $L$, $\tau$, and $\beta$ are the hyperparameters of \proposed.
Figure~\ref{fig:framework} shows the overall framework of \proposed.

The segmentation result for a test instance is obtained by the DTW alignment with its sequential pseudo-label $\slabs=\hat{y}_*\cdot\sumfuc(\actmaps)$, where $\hat{y}_*=\mathbb{I}[\score{*} \geq \tau_{*}]$. 
Since the instance-level label is not given for the test input, we instead impose the predicted label $\hat{y}_*$ as a binary mask;
that is, we filter out normal instances based on the global anomaly score $\score{*}$.
Once the model is trained, $\tau_{*}$ is automatically determined by its optimal value that achieves the best instance-level classification performance on the validation set.

\section{Experiments}
\label{sec:exp}

\subsection{Experimental Settings}
\label{subsec:expsetting}
\smallsection{Dataset}
For our experiments, we use four real-world temporal datasets collected from a range of tasks for detecting anomalous events, including multivariate time-series (MTS) and surveillance videos (Table~\ref{tbl:dataset}).
We split the set of all the instances by 5:2:3 ratio into a training set, a validation set, and a test set.
Note that only the instance-level labels are given for the training and validation set.
\begin{itemize}
    
    \setlength\itemsep{-0.2em}
    
    \item \textbf{Electromyography Dataset\footnote{http://archive.ics.uci.edu/ml/datasets/EMG+data+for+gestures} (\emg)}~\cite{lobov2018latent}:
    The 8-channel myographic signals recorded by bracelets worn on subjects' forearms.
    Among several types of gestures, ulnar deviation is considered as anomalous events.
    Each time-series instance contains the signals for 5 seconds, which are downsampled to 500 points.
    
    \item \textbf{Gasoil Plant Heating Loop Dataset\footnote{https://kas.pr/ics-research/dataset\_ghl\_1} (\ghl)}~\cite{filonov2016multivariate}: 
    The control sequences of a gasoil plant heating loop, which suffered cyber-attacks.
    As done in~\cite{wen2019time}, we crop 10 different instances of length 50,000 from each time-series, then downsample each of them to 1,000 points.
    
    \item \textbf{Server Machine Dataset\footnote{https://github.com/smallcowbaby/OmniAnomaly} (\smd)}~\cite{su2019robust}: 
    The 38-metric multivariate time-series from server machines over 5 weeks, collected from an internet company.  
    We split them every 720 points (i.e., 12 hours) to build the set of time-series instances.
    
    \item \textbf{Subway Exit Dataset  (\subway)}~\cite{adam2008robust}: The surveillance video for a subway exit gate, where each anomalous event corresponds to the passenger walking towards a wrong direction.
    We extract the visual features of each frame by using the pre-trained ResNet-34~\cite{he2016deep}, and make a single video instance include 450 frames.
\end{itemize}

\smallsection{Baselines}
We compare the performance of \proposed with that of anomaly detection methods for temporal data, which are based on three different approaches.
\begin{itemize}
\setlength\itemsep{-0.2em}
    \item \textbf{Unsupervised learning}: VAE-based methods that compute the point-level anomaly scores from the reconstruction of temporal inputs --- \donut~\cite{xu2018unsupervised}, \lstmvae~\cite{park2018multimodal}, \lstmndt~\cite{hundman2018detecting}, and \omnianom~\cite{su2019robust}.
    
    \item \textbf{Semi-supervised learning}: The variants of the unsupervised methods, whose models are trained by using only normal instances in order to make them utilize given instance-level labels\footnote{These methods are categorized as semi-supervised learning~\cite{chandola2009anomaly} in that they only utilize densely-labeled (normal) points from normal instances.} ---  \donutpp, \lstmvaepp, \lstmndtpp, and \omnianompp.
    
    \item \textbf{Weakly supervised learning}: The multiple-instance learning method that can produce the anomaly prediction for each fixed-length segment --- \deepmil~\cite{sultani2018real}.
\end{itemize}

For fair comparisons, \deepmil employs the same model architecture with \proposed (i.e., \dicnn).
Moreover, we consider different numbers of the segments in a single instance, denoted by \deepmil-4, 8, 16.
In case of the video dataset (i.e., \subway), the additional comparison with other frame-level video anomaly detection methods~\cite{ionescu2017unmasking, ionescu2019detecting, liu2018classifier, pang2020self} is presented in the supplementary material.
\begin{table}[t]
    \caption{Statistics of real-world temporal datasets.}
    \label{tbl:dataset}
    \centering
    \resizebox{.99\linewidth}{!} {%
    \begin{tabular}{rcccc}
    \toprule
    Dataset & \emg & \ghl & \smd & \subwayexit \\ \midrule
    Data Type & MTS & MTS & MTS & Video \\
    \#Variables & 8 & 19 & 38 & 1024 \\
    \#Points (Train) & 211,892 & 240,000 & 354,200 & 32,051 \\
    \#Points (Valid) & 84,734 & 96,000 & 141,670 & 13,050 \\
    \#Points (Test) & 127,199 & 144,000 & 212,550 & 19,800 \\
    Anomaly Ratio (\%) & 5.97 & 0.49 & 4.16 & 2.60 \\
    Instance Length & 500 & 1,000 & 720 & 450 \\
    \bottomrule
    \end{tabular}
    }
\end{table}

\smallsection{Evaluation Metrics}
For quantitative evaluation of the anomaly detection (and segmentation) results, we measure the F1-score (denoted by F1) and the intersection over union (denoted by IoU) by using the point-level ground truth:
F1 = $\frac{2\times \text{Precision}\times\text{Recall}}{\text{Precision}+\text{Recall}}$ where Precision = $\frac{TP}{TP+FP}$ and Recall = $\frac{TP}{TP+FN}$, and IoU = $\frac{TP}{TP+FP+FN}$.
Several previous work~\cite{park2018multimodal, sultani2018real, xu2018unsupervised} mainly reported the area under the ROC curve (AUROC) as their evaluation metric, but it is known to produce misleading results especially for severely imbalanced classification with few samples of the minority class~\cite{saito2015precision}, such as anomaly detection tasks.

To compute the precision and recall, the weakly supervised methods (i.e., \deepmil and \proposed) can find the optimal anomaly threshold (applied to instance-level or segment-level scores) by utilizing the instance-level labels from the validation set.
However, in cases of the unsupervised and semi-supervised methods, they require point-level anomaly labels to tune the anomaly threshold (applied to point-level scores), thus we report their F1 and IoU by using the anomaly threshold that yields the best F1 among all possible thresholds (denoted by F1-best and IoU-best, respectively).
This can be interpreted as a measure of the discrimination power between normal and anomalous points.
For some of the baselines (i.e., \lstmndt and \omnianom) that have their own techniques to automatically determine the anomaly threshold, we additionally report their F1 and IoU by using the selected threshold.

\begin{table*}[t]
\caption{Performances of \proposed and other baselines in detecting anomalous points from real-world temporal data. All the results support statistically significant improvement of \proposed over the best baseline ($p \leq 0.05$ from the paired $t$-test).}
\label{tbl:mainresults}
\centering
\resizebox{0.99\linewidth}{!}{%
\begin{tabular}{rcccccccc}
    \toprule
    \multirow{2.5}{*}{Methods} & \multicolumn{4}{c}{\emg} & \multicolumn{4}{c}{\ghl}  \\ \cmidrule{2-9}
    & F1 & IoU & F1-best & IoU-best & F1 & IoU & F1-best & IoU-best \\ \midrule
    \donut~\cite{xu2018unsupervised} & - & - & 0.1748(0.002) & 0.0958(0.001) & - & - & 0.0363(0.013) & 0.0185(0.007) \\
    \lstmvae~\cite{park2018multimodal} & - & - & 0.1728(0.000) & 0.0946(0.000) & - & - & 0.0746(0.015) & 0.0388(0.008) \\
    \lstmndt~\cite{hundman2018detecting} & 0.1317(0.016) & 0.0705(0.009) & 0.1880(0.010) & 0.1199(0.001) & 0.0640(0.030) & 0.0332(0.016) & 0.1025(0.023) & 0.0541(0.013) \\
    \omnianom~\cite{su2019robust} & 0.1574(0.003) & 0.0854(0.002) & 0.1793(0.001) & 0.0985(0.001) & 0.0611(0.033) & 0.0318(0.018) & 0.0743(0.026) & 0.0387(0.014)  \\
    \midrule
    
    \donutpp~\cite{xu2018unsupervised} & - & - & 0.1784(0.001) & 0.0980(0.001) & - & - & 0.0850(0.036) & 0.0447(0.019) \\
    \lstmvaepp~\cite{park2018multimodal} & - & - & 0.1745(0.000) & 0.0956(0.000) & - & - & 0.0828(0.023) & 0.0433(0.013) \\
    \lstmndtpp~\cite{hundman2018detecting} & 0.1344(0.005) & 0.0720(0.003) & 0.1916(0.011) & 0.1201(0.002) & 0.0984(0.091) & 0.0538(0.053) & 0.1043(0.106) & 0.0578(0.064) \\
    \omnianompp~\cite{su2019robust} & 0.1536(0.002) & 0.0832(0.001) & 0.1807(0.002) & 0.0993(0.001) & 0.1209(0.101) & 0.0668(0.058) & 0.2096(0.143) & 0.1231(0.094) \\
    \midrule
    
    \deepmil-4~\cite{sultani2018real} & 0.4699(0.022) & 0.3073(0.019) & - & - & 0.0690(0.035) & 0.0359(0.098) & - & - \\
    \deepmil-8~\cite{sultani2018real} & 0.4317(0.029) & 0.2755(0.023) & - & - & 0.0571(0.023) & 0.0323(0.015) & - & - \\
    \deepmil-16~\cite{sultani2018real} & 0.3182(0.056) & 0.1902(0.039) & - & - & 0.1497(0.040) & 0.0813(0.023) & - & - \\
    \textbf{\proposed (ours)} & \textbf{0.5803(0.068)} & \textbf{0.4118(0.064)} & - & - & \textbf{0.2295(0.028)} & \textbf{0.1298(0.018)} & - & - \\  
    \bottomrule
    \toprule
    \multirow{2.5}{*}{Methods} & \multicolumn{4}{c}{\smd} & \multicolumn{4}{c}{\subwayexit} \\ \cmidrule{2-9}
    & F1 & IoU & F1-best & IoU-best & F1 & IoU & F1-best & IoU-best  \\ \midrule
    \donut~\cite{xu2018unsupervised} & - & - & 0.3206(0.011) & 0.1909(0.008) & - & - & 0.5080(0.018) & 0.3406(0.016) \\
    \lstmvae~\cite{park2018multimodal} & - & - & 0.2671(0.018) & 0.1542(0.012) & - & - & 0.5329(0.024) & 0.3635(0.022)  \\
    \lstmndt~\cite{hundman2018detecting} & 0.1145(0.018) & 0.0608(0.010) & 0.3588(0.019) & 0.2187(0.014) & 0.0000(0.000) & 0.0000(0.000) & 0.5658(0.005) & 0.3945(0.005) \\
    \omnianom~\cite{su2019robust} & 0.1176(0.002) & 0.0625(0.001) & 0.1223(0.006) & 0.0651(0.004) & 0.5367(0.043) & 0.3704(0.038) & 0.6065(0.020) & 0.4355(0.020) \\
    \midrule
    
    \donutpp~\cite{xu2018unsupervised} & - & - & 0.2875(0.067) & 0.1693(0.047) & - & - & 0.5100(0.026) & 0.3449(0.021) \\
    \lstmvaepp~\cite{park2018multimodal} & - & - & 0.2477(0.015) & 0.1414(0.010) & - & - & 0.5452(0.016) & 0.3749(0.016) \\
    \lstmndtpp~\cite{hundman2018detecting} & 0.1211(0.010) & 0.0645(0.006) & 0.3819(0.020) & 0.2361(0.015) & 0.0000(0.000) & 0.0000(0.000) & 0.5723(0.004) & 0.4009(0.004) \\
    \omnianompp~\cite{su2019robust} & 0.1435(0.081) & 0.0790(0.049) & 0.1750(0.077) & 0.0974(0.047) & 0.5479(0.028) & 0.3790(0.026) & 0.6198(0.023) & 0.4494(0.024) \\
    \midrule
    
    \deepmil-4~\cite{sultani2018real} & 0.3561(0.052) & 0.2176(0.038) & - & - & 0.5138(0.081) & 0.3738(0.073) & - & - \\
    \deepmil-8~\cite{sultani2018real} & 0.3450(0.032) & 0.2088(0.023) & - & - & 0.6471(0.066) & 0.4885(0.064) & - & - \\
    \deepmil-16~\cite{sultani2018real} & 0.3568(0.016) & 0.2173(0.012) & - & - & 0.6102(0.077) & 0.4391(0.072) & - & -\\
    \textbf{\proposed (ours)} & \textbf{0.4358(0.046)} & \textbf{0.2795(0.037)} & - & - & \textbf{0.7414(0.023)} & \textbf{0.5907(0.028)}  & - & - \\  
    \bottomrule
\end{tabular}
}
\end{table*}

\begin{table}[t]
\caption{F1-scores of \proposed that ablates each component, Dataset: \smd. C and R denote the ``corruption'' and ``random sampling'' for negative pseudo-labels, respectively.}
\label{tbl:ablation}
\small
\centering
\resizebox{0.99\linewidth}{!}{%
\begin{tabular}{PccPPPP}
    \toprule
    Model Arch. & $\closs$ & $\aloss$ & DTW Seg. & Global Pool. & Neg. Label & F1 \\\midrule
    \lstm 
    & \checkmark & \checkmark & \checkmark & MAX & C & 0.3685 \\ \midrule
    \multirow{6.5}{*}{\dicnn} 
    & \checkmark & - & - & MAX & C & 0.1225 \\
    & \checkmark & \checkmark & - & MAX & C & 0.1265 \\
    & \checkmark & - & \checkmark & MAX & C & 0.3384 \\ \cmidrule{2-7}
    & \checkmark & \checkmark & \checkmark & AVG & C & 0.3046 \\
    & \checkmark & \checkmark & \checkmark & MAX & R & 0.4272 \\
    & \checkmark & \checkmark & \checkmark & MAX & C & \textbf{0.4358} \\
    \bottomrule
\end{tabular}
}
\end{table}

\smallsection{Implementation Details}
We implement our \proposed and all the baselines using PyTorch, and train them with the Adam optimizer~\cite{kingma2014adam}.
For the unsupervised methods, we tune their hyperparameters in the ranges suggested by the previous work~\cite{su2019robust} that considered the same baselines.
In case of \vae-based methods, we set the size of a temporal context for each point (i.e., sliding window) to 128 (for MTS) and 16 (for video).
For \dicnn-based methods, we stack 7 (for MTS) and 4 (for video) layers of dilated convolutions with filter size 2 to keep the size of its receptive field (=$2^7$, $2^4$) the same with the others'.
The dimensionality $d$ of hidden (and output) vectors in \dicnn and the smoothing factor $\gamma$ of soft-DTW are set to 128 and 0.01, respectively.
We provide the in-depth sensitivity analysis on the hyperparameters (i.e., $L$, $\tau$, and $\beta$) in our supplementary material.

\subsection{Experimental Results}
\label{subsec:expresults}
\smallsection{Performances on Anomaly Detection}
\label{subsubsec:performance}
We first measure the detection performance of \proposed and the other baselines.
In this experiment, we do not consider the point-adjust approach~\cite{su2019robust,xu2018unsupervised} for our evaluation strategy:
if any point in a ground truth anomalous segment is detected as the anomaly, all points in the segment are considered to be correctly detected as the anomalies.\footnote{This approach skews the results by excessively increasing the true positives, i.e., the F1 and IoU are overestimated.}
We repeat to train each model five times using different random seeds, and report the averaged results with their standard deviations.

\begin{figure*}[t]
	\centering
	\includegraphics[width=\linewidth]{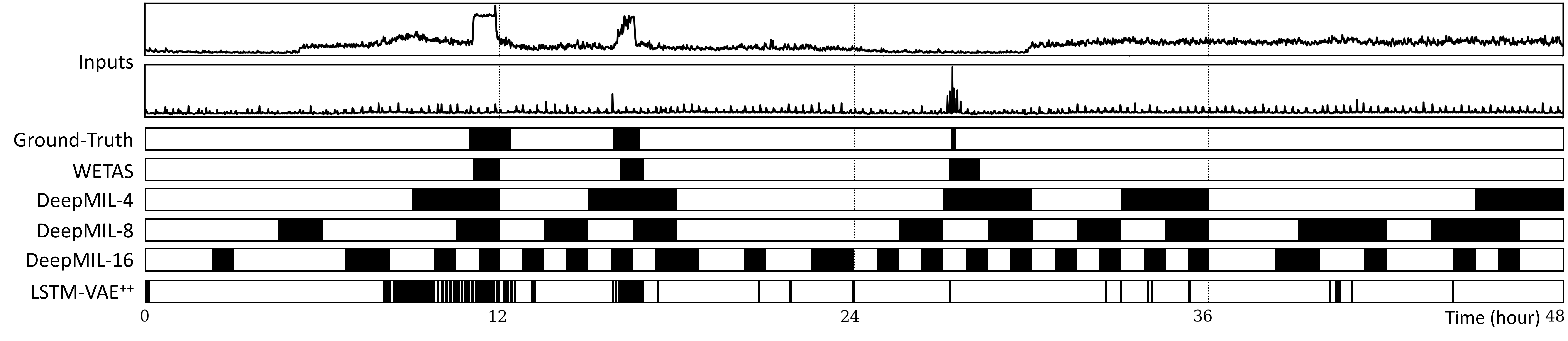}
	\vspace{-20pt}
	\caption{Anomaly detection results of \proposed and other baselines on 4 consecutive instances, Dataset: \smd. \proposed correctly detects anomalous points in terms of the number of ground truth anomalous segments and their boundaries. }
	\label{fig:qresults}
\end{figure*}

\begin{figure}[t]
	\centering		
	\includegraphics[width=\linewidth]{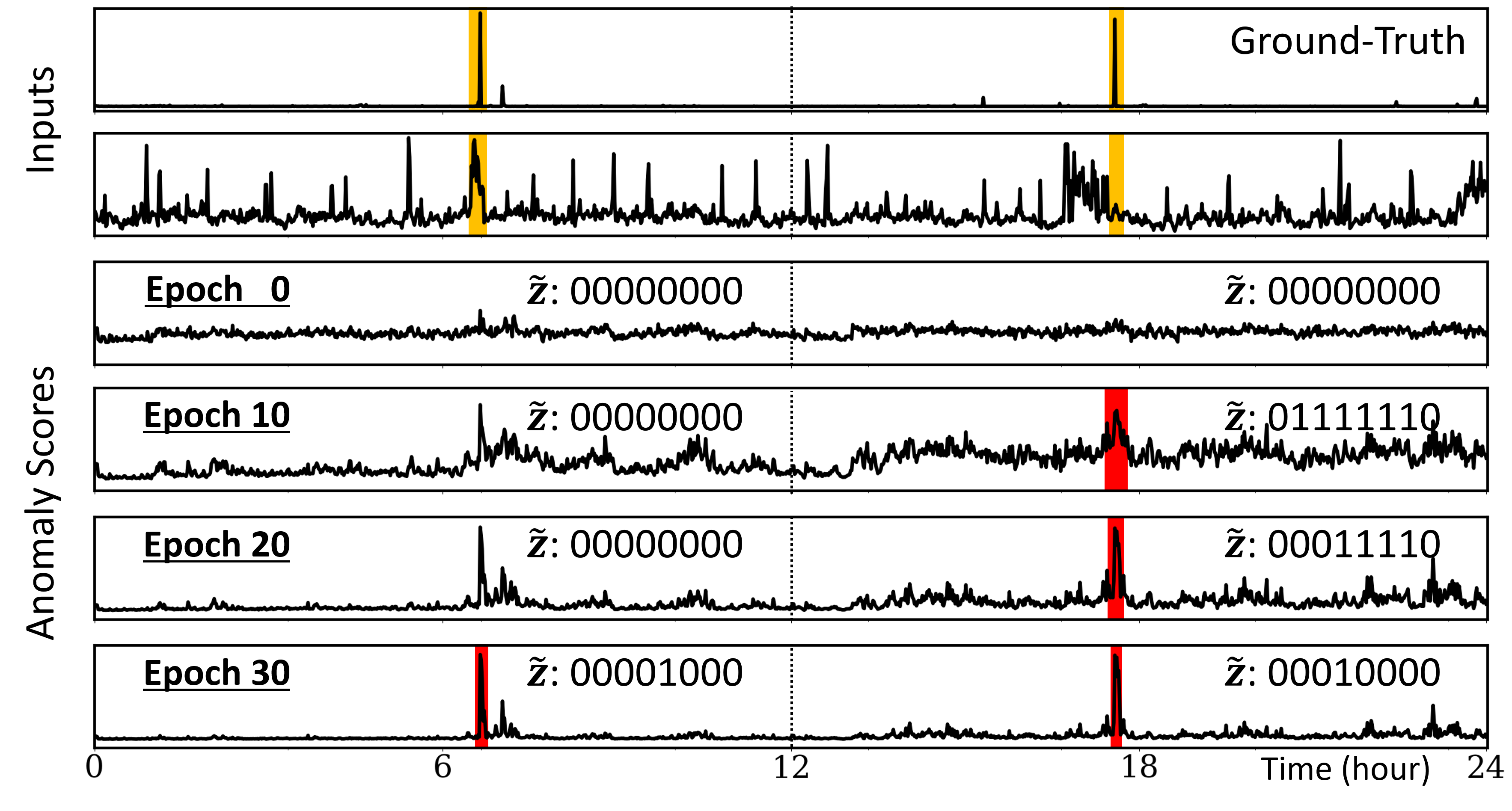}
	\vspace{-15pt}
	\caption{Test instances (upper) and the local anomaly scores computed by \proposed every 10 epoch (lower), Dataset: \smd. The ground truth and identified anomalous segments are colored in yellow and red, respectively.}
	\label{fig:trainingprocess}
\end{figure}

In Table~\ref{tbl:mainresults}, the unsupervised and semi-supervised methods show considerably worse performance than the weakly supervised methods in terms of both F1 and IoU.
Even considering F1-best and IoU-best, their performance is at most comparable to that of the weakly supervised methods.
In other words, though they additionally use the point-level labels for finding the best anomaly threshold that clearly distinguishes anomalous points from normal ones, they cannot achieve the detection performance as high as the weakly supervised methods due to the lack of supervision.

Compared to the unsupervised methods that use all training instances regardless of their instance-level labels, the semi-supervised methods that selectively use them (i.e., only normal instances) sometimes perform better and sometimes worse.
That is, learning the normality of temporal data from anomalous instances can degrade the detection performance due to anomalous patterns that reside in the instances;
but in some cases, the increasing number of training data points is rather helpful to improve the generalization power despite the existence of anomalous points.
This strongly indicates that how to utilize the instance-level labels does matter and largely affects the final performance.

Our \proposed achieves the best performance for all the datasets, and notably, it significantly beats \deepmil whose performance highly depends on the number of segments.
For the \emg dataset whose instance length is only 500, \deepmil-16 is not able to effectively detect anomalies because each input segment is too short to accurately compute the segment-level anomaly score;
for the \ghl dataset, \deepmil-4 suffers from coarse-grained segmentation, which leads to the limited F1 and IoU.
On the contrary, \proposed successfully finds anomalous segments by using dynamic alignment with the sequential pseudo-label.

\smallsection{Ablation Study}
\label{subsubsec:ablation}
To further investigate the effectiveness of \proposed, we measure the F1-score of several variants that ablate each of the following components:
1) \dicnn (vs. \lstm) for capturing the temporal dependency, 2) DTW-based alignment loss, 3) DTW-based segmentation, 
4) global max (vs. average) pooling for the classification loss, 
and 5) corruption by the instance-level label (vs. random sampling) for the negative pseudo-label $\slabs_-$ in Equation~\eqref{eq:aloss}.
In case of \proposed without the DTW-based segmentation, we measure both the F1\footnote{For this measure, we set the anomaly threshold (applied to our point-level anomaly scores) to $\tau$ that is used for pseudo-labeling in Equation~\eqref{eq:seqlabel}.} and F1-best using the series of local (or point-level) anomaly scores.


Table~\ref{tbl:ablation} summarizes the results of our ablation study on the \smd dataset.
It is worth noting that the DTW-based segmentation considerably improves the performance compared to the case simply using the obtained local anomaly scores.
This is because the alignment with the pseudo-label filters out normal instances based on the global anomaly score, and also considers the continuity of anomalous points while limiting the number of anomalous segments.
Nevertheless, their F1-best scores (for Rows 2 and 3) are respectively 0.4472 and 0.4590, which means that \proposed outperforms other point-level detection methods even without the DTW-based segmentation.
In addition, it is obvious that \dicnn is more effective to capture the long-term dependency of temporal data than \lstm.
The global max pooling and the corrupted pseudo-labels also turn out to help the \dicnn model to learn useful features from the instance-level labels, compared to their alternatives.

\smallsection{Qualitative Analyses}
\label{subsubsec:qualitative}
We also qualitatively compare the anomaly detection results
for test instances.
Figures~\ref{fig:qresults} and~\ref{fig:trainingprocess} present the results on consecutive instances for the \smd dataset.
We plot only the two series of input variables that are directly related to observed anomalous events.

Compared to the baselines, \proposed more accurately finds 
anomalous segments in terms of their number as well as the boundaries.
To be specific, \deepmil-4 with fewer segments correctly identifies the segments that include anomalous events, but the boundaries are far from the ground truth due to their fixed length and location.
In contrast, \deepmil-16 with more segments makes unreliable predictions, because each input segment is not long enough that the normality and abnormality of the segment are effectively captured.
The point-level detection method, \lstmvaepp, finds out non-consecutive anomalous points, which results in a large number of separate segments.
It is obvious that such discontinuous prediction makes it difficult to investigate the occurrence of anomalous events.

\section{Conclusion}
\label{sec:concl}
This paper proposes a weakly supervised learning framework for temporal anomaly segmentation, which effectively finds anomalous segments by leveraging instance-level anomaly labels for model training.
For each input instance of temporal data, the learning objective of \proposed includes its accurate classification as the instance-level label and also strong alignment with the sequential pseudo-label.
In the end, \proposed utilizes the DTW alignment between an input instance and its pseudo-label to obtain the temporal segmentation result.
Our empirical evaluation shows that \proposed outperforms all baselines in detecting temporal anomalies as well as specifying precise segments that correspond to anomalous events.

\smallsection{Acknowledgement}
This work was supported by the NRF grant (No. 2020R1A2B5B03097210) and the IITP grant (No. 2018-0-00584, 2019-0-01906) funded by the MSIT.

\balance
{\small
\bibliographystyle{ieee_fullname}
\bibliography{BIB/bibliography}

\begin{thebibliography}{10}\itemsep=-1pt

\bibitem{abdallah2016fraud}
Aisha Abdallah, Mohd~Aizaini Maarof, and Anazida Zainal.
\newblock Fraud detection system: A survey.
\newblock {\em Journal of Network and Computer Applications}, 68:90--113, 2016.

\bibitem{adam2008robust}
Amit Adam, Ehud Rivlin, Ilan Shimshoni, and Daviv Reinitz.
\newblock Robust real-time unusual event detection using multiple
  fixed-location monitors.
\newblock {\em TPAMI}, 30(3):555--560, 2008.

\bibitem{ahmad2017unsupervised}
Subutai Ahmad, Alexander Lavin, Scott Purdy, and Zuha Agha.
\newblock Unsupervised real-time anomaly detection for streaming data.
\newblock {\em Neurocomputing}, 262:134 -- 147, 2017.

\bibitem{bergmann2019mvtec}
Paul Bergmann, Michael Fauser, David Sattlegger, and Carsten Steger.
\newblock Mvtec ad--a comprehensive real-world dataset for unsupervised anomaly
  detection.
\newblock In {\em CVPR}, pages 9592--9600, 2019.

\bibitem{bergmann2020uninformed}
Paul Bergmann, Michael Fauser, David Sattlegger, and Carsten Steger.
\newblock Uninformed students: Student-teacher anomaly detection with
  discriminative latent embeddings.
\newblock In {\em CVPR}, pages 4183--4192, 2020.

\bibitem{chandola2009anomaly}
Varun Chandola, Arindam Banerjee, and Vipin Kumar.
\newblock Anomaly detection: A survey.
\newblock {\em ACM computing surveys}, 41(3):1--58, 2009.

\bibitem{chang2019d3tw}
Chien-Yi Chang, De-An Huang, Yanan Sui, Li Fei-Fei, and Juan~Carlos Niebles.
\newblock D3tw: Discriminative differentiable dynamic time warping for weakly
  supervised action alignment and segmentation.
\newblock In {\em CVPR}, pages 3546--3555, 2019.

\bibitem{cuturi2017soft}
Marco Cuturi and Mathieu Blondel.
\newblock Soft-dtw: a differentiable loss function for time-series.
\newblock In {\em ICML}, pages 894--903, 2017.

\bibitem{cuturi2007kernel}
Marco Cuturi, Jean-Philippe Vert, Oystein Birkenes, and Tomoko Matsui.
\newblock A kernel for time series based on global alignments.
\newblock In {\em ICASSP}, volume~2, pages II--413, 2007.

\bibitem{filonov2016multivariate}
Pavel Filonov, Andrey Lavrentyev, and Artem Vorontsov.
\newblock Multivariate industrial time series with cyber-attack simulation:
  Fault detection using an lstm-based predictive data model.
\newblock {\em arXiv preprint arXiv:1612.06676}, 2016.

\bibitem{gornitz2013toward}
Nico G{\"o}rnitz, Marius Kloft, Konrad Rieck, and Ulf Brefeld.
\newblock Toward supervised anomaly detection.
\newblock {\em JAIR}, 46:235--262, 2013.

\bibitem{hasan2016learning}
Mahmudul Hasan, Jonghyun Choi, Jan Neumann, Amit~K Roy-Chowdhury, and Larry~S
  Davis.
\newblock Learning temporal regularity in video sequences.
\newblock In {\em CVPR}, pages 733--742, 2016.

\bibitem{he2016deep}
Kaiming He, Xiangyu Zhang, Shaoqing Ren, and Jian Sun.
\newblock Deep residual learning for image recognition.
\newblock In {\em CVPR}, pages 770--778, 2016.

\bibitem{hendrycks2019benchmark}
Dan Hendrycks, Steven Basart, Mantas Mazeika, Mohammadreza Mostajabi, Jacob
  Steinhardt, and Dawn Song.
\newblock A benchmark for anomaly segmentation.
\newblock {\em arXiv preprint arXiv:1911.11132}, 2019.

\bibitem{hundman2018detecting}
Kyle Hundman, Valentino Constantinou, Christopher Laporte, Ian Colwell, and Tom
  Soderstrom.
\newblock Detecting spacecraft anomalies using lstms and nonparametric dynamic
  thresholding.
\newblock In {\em KDD}, pages 387--395, 2018.

\bibitem{ionescu2017unmasking}
Radu~Tudor Ionescu, Sorina Smeureanu, Bogdan Alexe, and Marius Popescu.
\newblock Unmasking the abnormal events in video.
\newblock In {\em ICCV}, pages 2895--2903, 2017.

\bibitem{tudor2017unmasking}
Radu~Tudor Ionescu, Sorina Smeureanu, Bogdan Alexe, and Marius Popescu.
\newblock Unmasking the abnormal events in video.
\newblock In {\em ICCV}, pages 2895--2903, 2017.

\bibitem{ionescu2019detecting}
Radu~Tudor Ionescu, Sorina Smeureanu, Marius Popescu, and Bogdan Alexe.
\newblock Detecting abnormal events in video using narrowed normality clusters.
\newblock In {\em WACV}, pages 1951--1960, 2019.

\bibitem{ju2020pumad}
Hyunjun Ju, Dongha Lee, Junyoung Hwang, Junghyun Namkung, and Hwanjo Yu.
\newblock Pumad: Pu metric learning for anomaly detection.
\newblock {\em Information Sciences}, 523:167--183, 2020.

\bibitem{kingma2014adam}
Diederik~P Kingma and Jimmy Ba.
\newblock Adam: A method for stochastic optimization.
\newblock {\em arXiv preprint arXiv:1412.6980}, 2014.

\bibitem{lee2021learnable}
Dongha Lee, Seonghyeon Lee, and Hwanjo Yu.
\newblock Learnable dynamic temporal pooling for time series classification.
\newblock In {\em AAAI}, volume~35, pages 8288--8296, 2021.

\bibitem{lee2019ficklenet}
Jungbeom Lee, Eunji Kim, Sungmin Lee, Jangho Lee, and Sungroh Yoon.
\newblock Ficklenet: Weakly and semi-supervised semantic image segmentation
  using stochastic inference.
\newblock In {\em CVPR}, pages 5267--5276, 2019.

\bibitem{liu2018classifier}
Yusha Liu, Chun-Liang Li, and Barnab{\'a}s P{\'o}czos.
\newblock Classifier two sample test for video anomaly detections.
\newblock In {\em BMVC}, page~71, 2018.

\bibitem{lobov2018latent}
Sergey Lobov, Nadia Krilova, Innokentiy Kastalskiy, Victor Kazantsev, and
  Valeri~A Makarov.
\newblock Latent factors limiting the performance of semg-interfaces.
\newblock {\em Sensors}, 18(4):1122, 2018.

\bibitem{long2015fully}
Jonathan Long, Evan Shelhamer, and Trevor Darrell.
\newblock Fully convolutional networks for semantic segmentation.
\newblock In {\em CVPR}, pages 3431--3440, 2015.

\bibitem{luo2017remembering}
Weixin Luo, Wen Liu, and Shenghua Gao.
\newblock Remembering history with convolutional lstm for anomaly detection.
\newblock In {\em ICME}, pages 439--444, 2017.

\bibitem{munir2018deepant}
Mohsin Munir, Shoaib~Ahmed Siddiqui, Andreas Dengel, and Sheraz Ahmed.
\newblock Deepant: A deep learning approach for unsupervised anomaly detection
  in time series.
\newblock {\em IEEE Access}, 7:1991--2005, 2018.

\bibitem{oord2016wavenet}
Aaron van~den Oord, Sander Dieleman, Heiga Zen, Karen Simonyan, Oriol Vinyals,
  Alex Graves, Nal Kalchbrenner, Andrew Senior, and Koray Kavukcuoglu.
\newblock Wavenet: A generative model for raw audio.
\newblock {\em arXiv preprint arXiv:1609.03499}, 2016.

\bibitem{pang2019deep}
Guansong Pang, Chunhua Shen, and Anton van~den Hengel.
\newblock Deep anomaly detection with deviation networks.
\newblock In {\em KDD}, pages 353--362, 2019.

\bibitem{pang2020self}
Guansong Pang, Cheng Yan, Chunhua Shen, Anton van~den Hengel, and Xiao Bai.
\newblock Self-trained deep ordinal regression for end-to-end video anomaly
  detection.
\newblock In {\em CVPR}, pages 12173--12182, 2020.

\bibitem{park2018multimodal}
Daehyung Park, Yuuna Hoshi, and Charles~C Kemp.
\newblock A multimodal anomaly detector for robot-assisted feeding using an
  lstm-based variational autoencoder.
\newblock {\em IEEE Robotics and Automation Letters}, 3(3):1544--1551, 2018.

\bibitem{richard2018neuralnetwork}
Alexander Richard, Hilde Kuehne, Ahsan Iqbal, and Juergen Gall.
\newblock Neuralnetwork-viterbi: A framework for weakly supervised video
  learning.
\newblock In {\em CVPR}, pages 7386--7395, 2018.

\bibitem{ronneberger2015u}
Olaf Ronneberger, Philipp Fischer, and Thomas Brox.
\newblock U-net: Convolutional networks for biomedical image segmentation.
\newblock In {\em MICCAI}, pages 234--241, 2015.

\bibitem{saito2015precision}
Takaya Saito and Marc Rehmsmeier.
\newblock The precision-recall plot is more informative than the roc plot when
  evaluating binary classifiers on imbalanced datasets.
\newblock {\em PloS one}, 10(3):e0118432, 2015.

\bibitem{sakoe1978dynamic}
Hiroaki Sakoe and Seibi Chiba.
\newblock Dynamic programming algorithm optimization for spoken word
  recognition.
\newblock {\em IEEE transactions on acoustics, speech, and signal processing},
  26(1):43--49, 1978.

\bibitem{su2019robust}
Ya Su, Youjian Zhao, Chenhao Niu, Rong Liu, Wei Sun, and Dan Pei.
\newblock Robust anomaly detection for multivariate time series through
  stochastic recurrent neural network.
\newblock In {\em KDD}, pages 2828--2837, 2019.

\bibitem{sultani2018real}
Waqas Sultani, Chen Chen, and Mubarak Shah.
\newblock Real-world anomaly detection in surveillance videos.
\newblock In {\em CVPR}, pages 6479--6488, 2018.

\bibitem{wang2017time}
Zhiguang Wang, Weizhong Yan, and Tim Oates.
\newblock Time series classification from scratch with deep neural networks: A
  strong baseline.
\newblock In {\em IJCNN}, pages 1578--1585, 2017.

\bibitem{wei2018anomaly}
Qi Wei, Yinhao Ren, Rui Hou, Bibo Shi, Joseph~Y Lo, and Lawrence Carin.
\newblock Anomaly detection for medical images based on a one-class
  classification.
\newblock In {\em Medical Imaging 2018: Computer-Aided Diagnosis}, volume
  10575, page 105751M, 2018.

\bibitem{wen2019time}
Tailai Wen and Roy Keyes.
\newblock Time series anomaly detection using convolutional neural networks and
  transfer learning.
\newblock {\em arXiv preprint arXiv:1905.13628}, 2019.

\bibitem{xu2018unsupervised}
Haowen Xu, Wenxiao Chen, Nengwen Zhao, Zeyan Li, Jiahao Bu, Zhihan Li, Ying
  Liu, Youjian Zhao, Dan Pei, Yang Feng, et~al.
\newblock Unsupervised anomaly detection via variational auto-encoder for
  seasonal kpis in web applications.
\newblock In {\em WWW}, pages 187--196, 2018.

\bibitem{yu2019temporal}
Tan Yu, Zhou Ren, Yuncheng Li, Enxu Yan, Ning Xu, and Junsong Yuan.
\newblock Temporal structure mining for weakly supervised action detection.
\newblock In {\em ICCV}, pages 5522--5531, 2019.

\bibitem{yuan2017temporal}
Zehuan Yuan, Jonathan~C Stroud, Tong Lu, and Jia Deng.
\newblock Temporal action localization by structured maximal sums.
\newblock In {\em CVPR}, pages 3684--3692, 2017.

\bibitem{zhang2018human}
Yong Zhang, Yu Zhang, Zhao Zhang, Jie Bao, and Yunpeng Song.
\newblock Human activity recognition based on time series analysis using u-net.
\newblock {\em arXiv preprint arXiv:1809.08113}, 2018.

\end{thebibliography}
}

\newpage
\appendix

\section{Pseudo-code of \proposed Framework}

Algorithm~\ref{alg:pseudocode} shows the pseudo-code of training the framework.
In practice, the model parameters are updated by using minibatch SGD with the Adam optimizer~\cite{kingma2014adam}.

\begin{algorithm}[h]
	\small
    \DontPrintSemicolon
    \SetKwComment{Comment}{$\triangleright$\ }{} 
	\KwIn{A training set containing $N$ instances of temporal data $\{(\vect^{(1)},y^{(1)}),\ldots,(\vect^{(N)},y^{(N)})\}$} 
	\KwOut{Updated \dicnn model $f(\cdot;\Theta)$ and the anomaly weight vector $\weightg$}
	\While{convergence}{
    \For{$i=1,\ldots,N$}{
    \vspace{5pt}
    \Comment*[r]{Compute the local anomaly scores}
    $\vech{1}^{(i)},\ldots,\vech{T}^{(i)} = f(\vecx{1}^{(i)},\ldots,\vecx{T}^{(i)};\Theta)$ \;
    $s_1^{(i)},\ldots,s_T^{(i)} = \sigma(\weightl^\top\vech{1}^{(i)}),\ldots,\sigma(\weightl^\top\vech{T}^{(i)})$ \;
 
    \vspace{5pt}
    \Comment*[r]{Compute the global anomaly score}
    $\vech{*}^{(i)} = \text{global-pooling}(\vech{1}^{(i)},\ldots,\vech{T}^{(i)})$ \;
    $s_*^{(i)} = \sigma(\weightg^\top \vech{*}^{(i)})$\;

    \vspace{5pt}
    \Comment*[r]{Obtain the sequential pseudo-label}
    $m_1^{(i)},\ldots,m_T^{(i)} = $ \; 
    \nonl\hspace{30pt} $\text{min-max-norm}(\weightg^\top\vech{1}^{(i)},\ldots,\weightg^\top\vech{T}^{(i)})$\;
    $\slabs_+^{(i)} = y^{(i)}\cdot\phi_L(m_1^{(i)},\ldots,m_T^{(i)})$\;
    $\slabs_-^{(i)} = (1-y^{(i)})\cdot\phi_L(m_1^{(i)},\ldots,m_T^{(i)})$\;

    \vspace{5pt}
    \Comment*[r]{Calculate the two losses}
    $\closs = -y^{(i)}\log s_*^{(i)} - (1-y^{(i)})\cdot\log(1-s_*^{(i)})$ \;
    $\aloss = \max(0, (1/T)\cdot\sdtw{\gamma}(\slabs_+^{(i)}, \vect^{(i)})$ \;
    \nonl\hspace{44pt} $- (1/T)\cdot\sdtw{\gamma}(\slabs_-^{(i)}, \vect^{(i)}) +\beta)$ \;
    $\mathcal{L} = \closs + \aloss$ \;

    \vspace{5pt}
    \Comment*[r]{Update all the model parameters}
    $\Theta = \Theta - \eta\cdot {\partial\mathcal{L}}/{\partial\Theta}$\;
    $\weightg = \weightg - \eta\cdot {\partial\mathcal{L}}/{\partial\weightg}$\;
    }}
\caption{Optimizing the \proposed framework}
\label{alg:pseudocode}
\end{algorithm}

\section{Computation of Dynamic Time Warping}

The dynamic time warping (DTW) discrepancy (and the optimal alignment matrix) between two time-series of lengths $M$ and $N$ is usually computed by solving a dynamic program based on Bellman recursion, which takes a quadratic $O(MN)$ cost.
The continuous relaxation of DTW (i.e., soft-DTW), which enables to calculate the gradient of DTW with respect to its input, also can be computed in a similar way to the original DTW~\cite{cuturi2017soft}. 
Algorithm~\ref{alg:softdtw} presents the detailed algorithms for computing $\sdtw{\gamma}(\slabs, \vect)$ and $\nabla_{\Delta(\slabs, \vect)}\sdtw{\gamma}(\slabs, \vect)$ based on Bellman recursion.

In order to obtain the eligible segmentation result, we need to enforce the constraint that a single time point should not be aligned with multiple consecutive labels.
To this end, our forward recursion which computes $\sdtw{\gamma}(\slabs, \vect)$ does not consider the $\downarrow$ relation in its recurrence; i.e., $R_{l,t}$ depends on only $R_{l-1,t-1}$ and $R_{l,t-1}$, excluding $R_{l-1,t}$ (Line 6 in Algorithm~\ref{alg:softdtw}).
Accordingly, the backward recursion which computes $\nabla_{\Delta(\slabs, \vect)}\sdtw{\gamma}(\slabs, \vect)$ also does not allow the $\uparrow$ relation; i.e., $E_{l,t}$ is obtained from $E_{l,t+1}$ and $E_{l+1, t+1}$, excluding $E_{l+1, t}$ (Line 16 in Algorithm~\ref{alg:softdtw}).

Note that the gradient of soft-DTW with respect to the cost matrix, $\nabla_{\Delta(\slabs, \vect)}\sdtw{\gamma}(\slabs, \vect)$, can effectively update the anomaly weight vector $\weightg$ as well as the model parameters of \dicnn $\Theta$ by the help of the gradient back-propagation.
This is possible because each entry of the cost matrix $[\Delta(\slabs, \vect)]_{lt} = \delta(\slab{l}, \vecx{t})$ is defined by the binary cross entropy between a pseudo-label $\slab{l}$ and a local anomaly score $s_t=\sigma(\weightg^\top \vech{t})$,
\begin{equation*}
    \delta(\slab{l}, \vecx{t}) = -\left\{\slab{l}\cdot\log s_t + (1-\slab{l})\cdot\log(1-s_t)\right\}.
\end{equation*}
In the end, the gradient optimizes each local anomaly score to be closer to the pseudo-labels that are softly-aligned with the time point.

\begin{algorithm}[h]
\small
    \DontPrintSemicolon
    \SetKwProg{Fn}{Function}{:}{}
    \SetKwComment{Comment}{$\triangleright$\ }{} 
    \Fn{$\text{forward} \left(\slabs, \vect \right)$}{
    \Comment*[r]{Fill the alignment cost matrix $R\in\mathbb{R}^{L\times T}$}
    $R_{0,0} = 0$ \;
    $R_{:,0} = R_{0,:} = \infty$ \; 
    \For{$l=1,\ldots,L$}{
        \For{$t=1,\ldots,T$}{
        \vspace{2pt}
            \small
            $R_{l,t} = \delta(\slab{l}, \vecx{t}) + \text{min}_\gamma\{R_{l-1,t-1}, R_{l,t-1} \}$
        }
    }
    \Return $\sdtw{\gamma}(\slabs, \vect) = R_{L,T}$
    }
    \vspace{5pt}
    \Fn{$\text{backward} \left(\slabs, \vect \right)$}{
    \Comment*[r]{Fill the soft alignment matrix $E\in\mathbb{R}^{L\times T}$} 
    $E_{L+1,T+1} = 1$ \Comment*[r]{$E_{l,t}:=\partial R_{L,T}/\partial R_{l,t}$}
    $E_{:,T+1} = E_{L+1,:} = 0$ \;
    $R_{:,T+1} = R_{L+1,:} = -\infty$ \;
    \For{$l=L,\ldots,1$}{
        \For{$t=T,\ldots,1$}{
        \vspace{2pt}
        $a = \exp\frac{1}{\gamma}(R_{l,t+1}-R_{l,t}-\delta(\slab{l},\vecx{t+1}))$ \;
        $b = \exp\frac{1}{\gamma}(R_{l+1,t+1}-R_{l,t}-\delta(\slab{l+1},\vecx{t+1}))$ \;
        \vspace{-10pt}
        $E_{l,t} = a \cdot E_{l,t+1} + b \cdot E_{l+1,t+1}$ \;
        }
    }
    \Return $\nabla_{\Delta(\slabs, \vect)}\sdtw{\gamma}(\slabs, \vect) = E$
    }
	\caption{Forward recursion and backward recursion to compute $\sdtw{\gamma}(\slabs, \vect)$ and  $\nabla_{\Delta(\slabs, \vect)}\sdtw{\gamma}(\slabs, \vect)$}
	\label{alg:softdtw}
\end{algorithm}

\smallsection{Complexity Analysis}
In terms of efficiency, we analyze that our framework additionally takes the computational cost of $O(LT)$ for DTW alignment between $\widetilde{\mathbf{z}}$ and $\mathbf{X}$ (per CNN inference and its gradient back-propagation), as described in Algorithm~\ref{alg:softdtw}.
Furthermore, because of 1) the small $L$ value ($L \ll T$), 2) batch-wise computations in the PyTorch framework, and 3) GPU parallel computations for DTW recursions based on the numba library,
it does not raise any severe efficiency issue.

\section{Reproducibility}
For reproducibility, our implementation is publicly available\footnote{https://github.com/donalee/wetas}, and Table~\ref{tbl:optim} presents the optimization details of \proposed.
We empirically found that the performances are hardly affected by these hyperparameters for the optimization.

\begin{table}[htbp]
    \caption{Details for the optimization of \proposed.}
    \label{tbl:optim}
    \centering
    \resizebox{.99\linewidth}{!} {%
    \begin{tabular}{rl}
    \toprule
    \multirow{2}{*}{Batch size} & 32 (for \emg, \ghl, \smd) \\
    & \ \ 8 (for \subway) \\
    Optimizer & Adam optimizer  \\
    Initial learning rate & 0.0001 \\
    Max \# epochs & 200 \\
    Stopping criterion & Instance-level F1 (on validation) \\ \bottomrule
    \end{tabular}
    }
\end{table}

\section{Hyperparaemter Search}

For our \proposed framework, we search the optimal values of the following hyperparameters: 
the length of a sequential pseudo-label $L\in$ \{4, 8, 12, 16\}, the anomaly threshold for pseudo-labeling $\tau\in$ \{0.1, 0.3, 0.5, 0.7\}, and the margin size for the alignment loss $\beta\in$ \{0.1, 0.5, 1.0, 2.0\}.
The optimal values are selected based on instance-level F1-scores on the validation set.
The selected hyperparameter values for reporting the final performance are listed in Table~\ref{tbl:hyperparameters}.
In case of the smoothing parameter $\gamma$ used for soft-DTW, we fix its value to 0.1 without further tuning.

\begin{table}[htbp]
    \caption{Selected hyperparameter values for \proposed.}
    \label{tbl:hyperparameters}
    \centering
    \begin{tabular}{rHHHH}
    \toprule
    Datasets & \emg  & \ghl & \smd & \subway  \\ \midrule
    global-pooling & avg & max & max & avg  \\
    $L$ & 4 & 16 & 12 & 12 \\
    $\tau$ & 0.3 & 0.1 & 0.5 & 0.5 \\
    $\beta$ & 0.1 & 1.0 & 0.5 & 0.5 \\ \bottomrule
    \end{tabular}
\end{table}

\section{Baseline Methods}
We describe the details of the anomaly detection methods which are used as the baseline in our experiments.
All of them employ their own deep neural networks to effectively model the temporal dependency among time points, and compute the anomaly score for each point or segment.
\begin{itemize} 
    \setlength\itemsep{-0.1em}
    \item \textbf{\donut}~\cite{xu2018unsupervised}:
    A simple VAE model optimized by the modified evidence lower bound (M-ELBO). 
    It also uses a sampling-based imputation technique for missing points, in order to effectively deal with anomalous points during the detection.
    
    \item \textbf{\lstmvae}~\cite{park2018multimodal}: 
    A VAE model that employs long short term memory (LSTM) for its encoder and decoder. 
    It is trained to reconstruct the non-anomalous training data well, and defines the anomaly score by the reconstruction error.
    
    \item \textbf{\lstmndt}~\cite{hundman2018detecting}:
     A LSTM network that is trained to predict the next input (i.e., sequence modeling). 
     It additionally adopts the non-parametric dynamic thresholding (NDT) technique to automatically determine the optimal anomaly threshold.
    
    \item \textbf{\omnianom}~\cite{su2019robust}: 
    The state-of-the-art point-level anomaly detection model that uses gated recurrent units (\gru) as the encoder and decoder of VAE. 
    It incorporates advanced techniques into VAE, including normalizing flows and a linear Gaussian space model, to consider stochasticity and temporal dependency among the time points.

    \item \textbf{\deepmil}~\cite{sultani2018real}: The multiple-instance learning method
    that learns from weakly labeled temporal data.
    It produces the anomaly prediction for each fixed-length segment, thus the result can be used for temporal anomaly segmentation.
    We consider different numbers of the segments in a single instance, denoted by \deepmil-4, 8, 16. 
    
\end{itemize}
Note that \donut, \lstmvae, \lstmndt, and \omnianom fall into the category of unsupervised learning as they do not utilize any anomaly labels for training, and their variants that use only normal instances are categorized as semi-supervised learning.
\deepmil is the only existing method that is based on weakly supervised learning.

\section{Additional Experiments}
\smallsection{Comparison with frame-level video anomaly detectors}
\label{subsec:videoperf}
In case of the video dataset (i.e., \subway), we additionally report the AUROC scores of unsupervised video anomaly detection methods~\cite{tudor2017unmasking, ionescu2019detecting, liu2018classifier, pang2020self} as a benchmark.
They aim to train the networks that take spatio-temporal (or spatial) inputs to compute the anomaly score of each video frame.
Note that all these methods produce the frame-level (or point-level) anomaly predictions, without utilizing the instance-level anomaly labels for training.
In Table~\ref{tbl:videoperf}, the weakly supervised methods\footnote{To compute AUROC for the weakly supervised methods, we regard the anomaly score of each segment as the score for all points in the segment (for \deepmil), and use the local (or point-level) anomaly scores without the DTW-based segmentation (for \proposed).}
(i.e., \deepmil and \proposed) show better performance than the unsupervised methods.
Even though \deepmil and \proposed simply use the pre-computed visual features of each frame (extracted by the pre-trained ResNet), it outperforms the other domain-specific baseline methods (optimized in an end-to-end manner) by the help of the instance-level anomaly labels.
This indicates that leveraging the weak supervision can be more effective to discriminate normal and anomalous video frames compared to fine-tuning the networks for visual feature extraction.

\begin{table}[t]
    \caption{Comparison with several recent frame-level video anomaly detection methods, Dataset: \subway.}
    \label{tbl:videoperf}
    \centering
    \resizebox{.99\linewidth}{!} {%
    \begin{tabular}{ABAB}
    \toprule
    \multicolumn{1}{c}{Methods} & AUROC & \multicolumn{1}{c}{Methods} & AUROC  \\ \midrule
    Ionescu et al.~\cite{tudor2017unmasking} & 85.7\% & \deepmil-4~\cite{sultani2018real} & 95.7\% \\
    Ionescu et al.~\cite{ionescu2019detecting} & 95.1\% & \deepmil-8~\cite{sultani2018real} & 96.9\% \\
    Liu et al.~\cite{liu2018classifier} & 93.1\% & \deepmil-16~\cite{sultani2018real} & 96.4\% \\
    Pang et al.~\cite{pang2020self} & 92.7\% & \proposed (ours) & 97.8\% \\ \bottomrule
    \end{tabular}
    }
\end{table}

\begin{figure}[t]
	\centering
	\includegraphics[width=\linewidth]{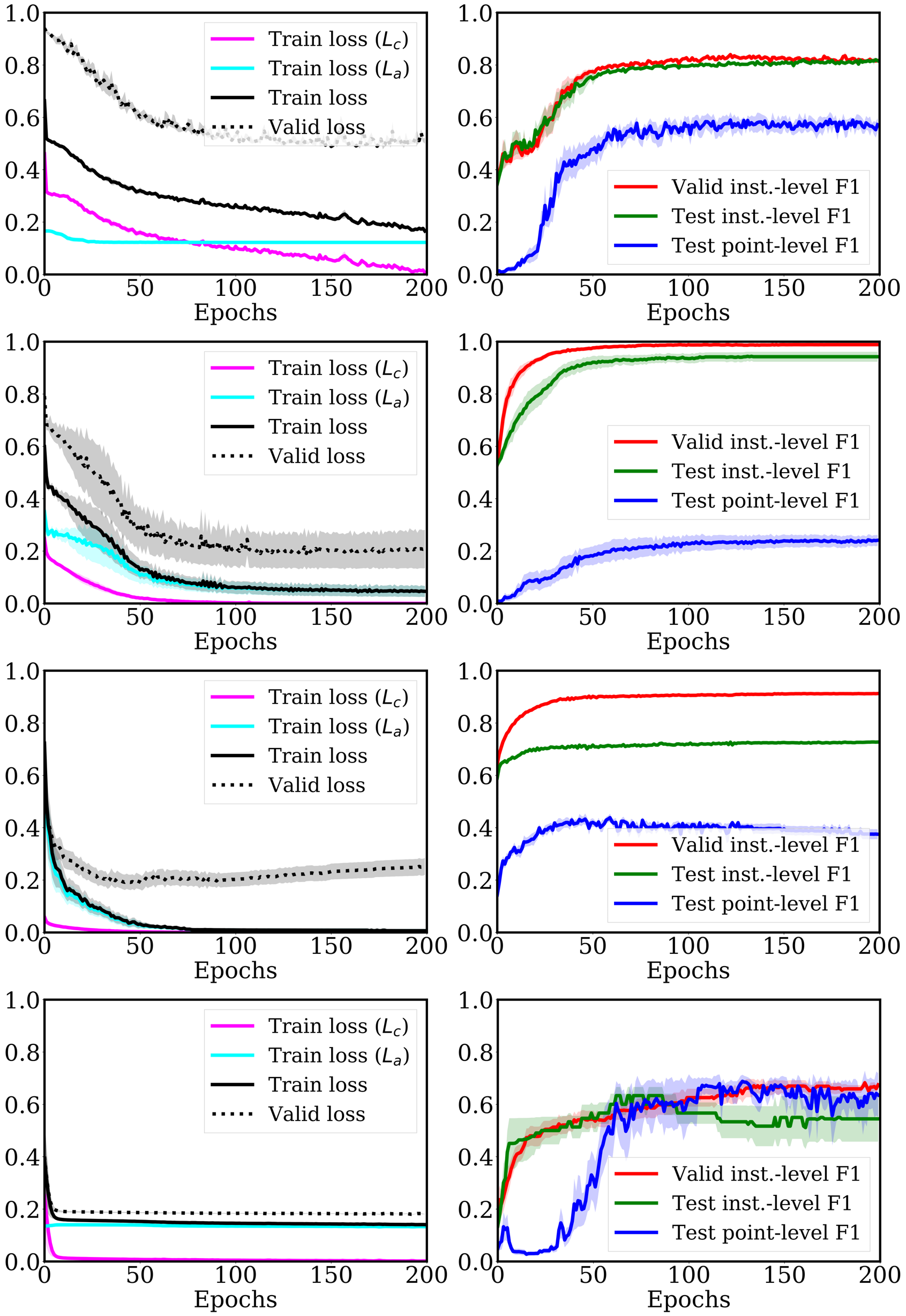}
	\caption{Learning curves of \proposed in terms of training/validation loss and test/validation F1-score, Datasets: \emg, \ghl, \smd, and \subway (from the first row).}
	\label{fig:traincurves}
\end{figure}

\smallsection{Learning curves}
\label{subsubsec:learncurve}
We plot the learning curves of \proposed by using the \emg, \ghl, \smd, and \subway datasets.
In Figure~\ref{fig:traincurves}, as the number of epochs increases, the classification loss and alignment loss consistently decrease for both the training and validation sets.
This implies that \proposed can infer more accurate sequential pseudo-labels as the training progresses, and the alignment loss better guides its model to output anomaly scores that are well aligned with the pseudo-label.
Similarly, Figure~\ref{fig:trainingprocess} illustrates that the pseudo-labeling and the DTW-based segmentation collaboratively improve with each other.
Consequently, the instance-level and point-level F1-scores for the test set increase as well.
The results empirically show that the instance-level F1-score (or the total loss) on the validation set can be a good termination criterion for the optimization of \proposed, thus we adopt it in our experiments.



\begin{figure}[h]
	\centering
	\includegraphics[width=0.99\linewidth]{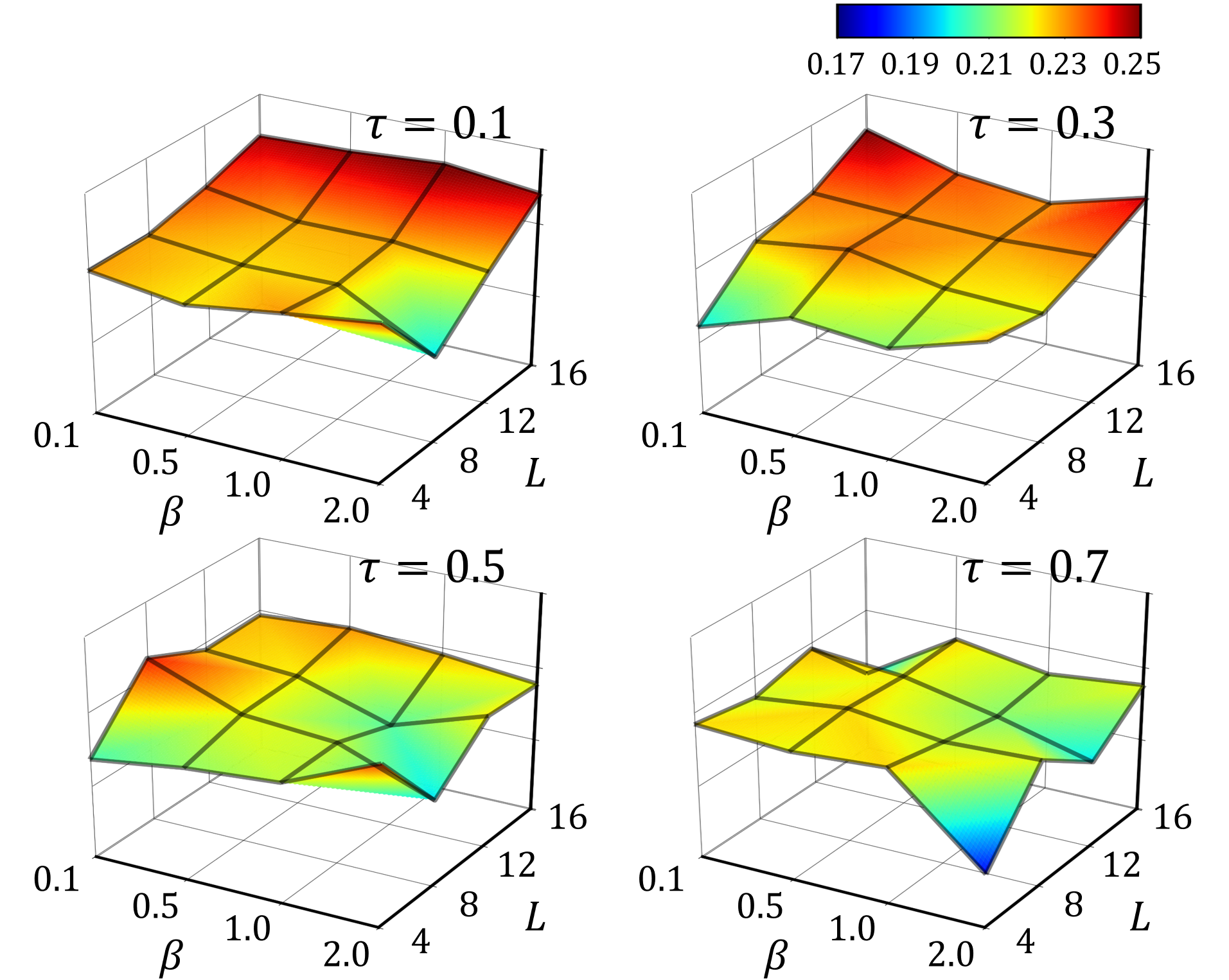}
	\vspace{10pt}
	\caption{F1-scores of \proposed with different hyperparameter values, Dataset: \ghl.}
	\label{fig:sensitivity}
\end{figure}

\smallsection{Sensitivity Analyses}
\label{subsubsec:sensitivity}
We finally examine the performance changes of \proposed with respect to the three hyperparameters (i.e., $L$, $\tau$, and $\beta$).
In Figure~\ref{fig:sensitivity}, we observe that the final performance of \proposed is not sensitive to $\beta$, and it shows higher F1-score when using a smaller $\tau$ and a larger $L$.
Specifically,
\begin{itemize}
    \item $\beta$ does not much affect the final performance of WETAS, because it simply controls the margin size in the alignment loss. 
    \item A smaller $\tau$ encourages to find out more anomalous segments, which leads to a high recall for anomaly detection, by making the sequential pseudo-label have more number of 1s.
    \item A larger $L$ allows a finer-grained segmentation by aligning the time points with more number of 0s and 1s in each sequential anomaly pseudo-label.
\end{itemize}
Nevertheless, unlike the \deepmil, the granularity of pseudo-labeling is not a critical factor for \proposed because the DTW-based segmentation is capable of dynamically aligning an input instance with its pseudo-label.
In conclusion, the best performing hyperparameter values successfully optimize \proposed under the weak supervision so that it can identify variable-length anomalous segments.

\end{document}